%% file: FADE.tex
\documentclass[letterpaper]{article} 
\usepackage{aaai2027}
\usepackage[hyphens]{url}  
\usepackage{graphicx} 
\usepackage{natbib}  
\usepackage{caption} 
\usepackage{algorithm}
\usepackage{algorithmic}

\usepackage{multirow}
\usepackage[table,xcdraw]{xcolor}
\usepackage{adjustbox}
\usepackage{newfloat}
\usepackage{amsmath}
\usepackage{amssymb}
\usepackage{amsfonts}
\usepackage{listings}
\usepackage{cleveref}

\DeclareCaptionStyle{ruled}{labelfont=normalfont,labelsep=colon,strut=off} 
\floatstyle{ruled}
\newfloat{listing}{tb}{lst}{}
\floatname{listing}{Listing}

\usepackage{booktabs}

\title{FADE: From Passive Verification to Active Discovery in Counterfactual Video Understanding}
\author{
    Fufangchen Zhao\textsuperscript{\rm 1 \rm 2}, Jinhu Fu\textsuperscript{\rm 1 \rm 2}, Jiachen Lei\textsuperscript{\rm 2}, Jiahong Wu\textsuperscript{\rm 2}\thanks{Project lead}, Xiangxiang Chu\textsuperscript{\rm 2}, Danfeng Yan\textsuperscript{\rm 1}\corresponding\\
}
\affiliations{
    \textsuperscript{\rm 1}State Key Laboratory of Networking and Switching Technology, BUPT\\

    \textsuperscript{\rm 2} AMAP, Alibaba Group\\
    zhaofufangchen@bupt.edu.cn
}

\makeatletter
\g@addto@macro\@maketitle{%
  \begingroup
    \centering
    \includegraphics[width=\textwidth]{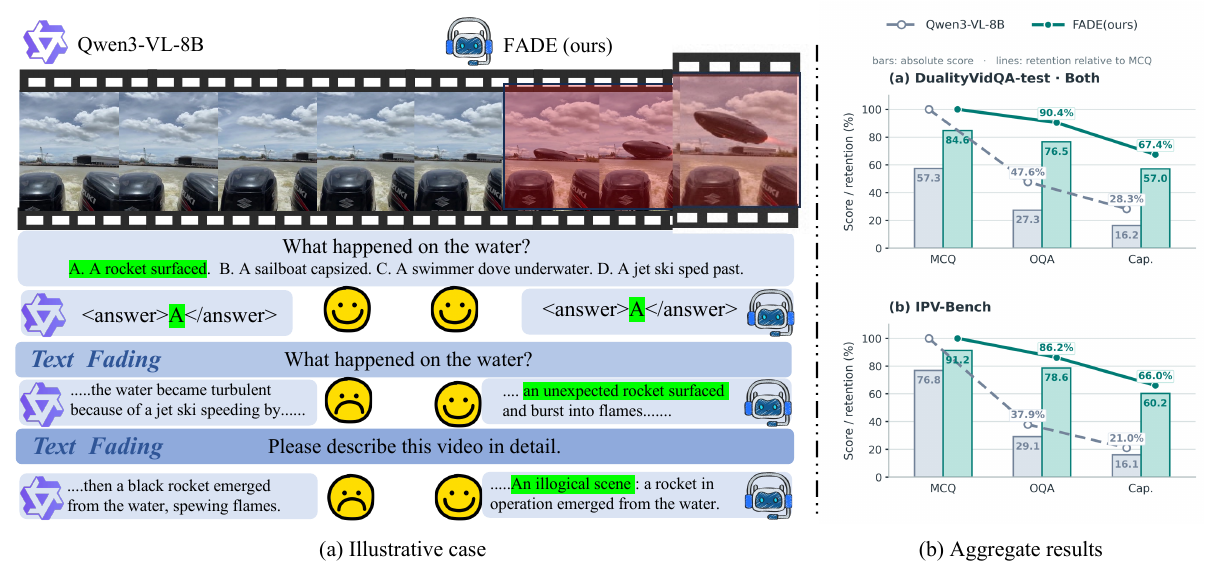}\par
    \captionof{figure}{
      \small Performance comparison between FADE and its base model as instance-specific textual anchors progressively fade. (a) Both models answer the anchor-rich MCQ correctly; after the options and then the question-specific cue are removed, only FADE consistently discovers and explains the counterfactual event. (b) Aggregate results on DualityVidQA-test and IPV-Bench. Bars report absolute scores, while lines report performance retained relative to MCQ.
    }
    \label{fig:teaser}
  \endgroup
}
\makeatother

\begin{document}

\maketitle

\begin{abstract}

\input{sections/0_abstract}
\end{abstract}


\input{sections/1_intro}

\input{sections/2_relate}
\input{sections/3_method}
\input{sections/4_experiement}
\input{sections/5_conclusion}

\bibliography{FADE}


\end{document}

%% file: sections/0_abstract.tex
Counterfactual video understanding evaluates whether models grasp physical and commonsense regularities. However, existing multiple-choice question (MCQ) benchmarks inadvertently leak target events through their questions and candidate options. This reduces the core challenge from active discovery to text-guided verification.

In this paper, we present FADE, an effective training framework for counterfactual discovery and explanation. Our method is built on an evidence-first, two-stage training paradigm. First, evidence-internalized supervised fine-tuning grounds the model's predictions in decisive visual anomalies. Second, we apply a fading-anchor reinforcement learning strategy that progressively removes textual guidance, compelling the model to independently discover and explain evidence. To rigorously evaluate this capability, we also introduce an effective pipeline that converts existing MCQ datasets into aligned MCQ, open-ended question answering (OQA), and captioning tasks without requiring additional data curation.

Our simple approach yields strong results. Using Qwen3-VL-8B as the baseline, FADE achieves state-of-the-art strict paired scores across all three tasks on DualityVidQA-test and IPV-Bench, outperforming GPT-5.6. 
In specific, when transitioning from constrained MCQs to unconstrained OQA and captioning, our model demonstrates remarkable robustness. Its performance retention is 90.4\% and 67.4\% on DualityVidQA-test—substantially higher than the 48.1\% and 30.7\% retained by GPT-5.6. We hope this simple framework can serve as a solid baseline for future research in unconstrained counterfactual video understanding.

%% file: sections/1_intro.tex
\section{Introduction}
\begin{figure*}[t]
    \centering
    \includegraphics[width=\textwidth]{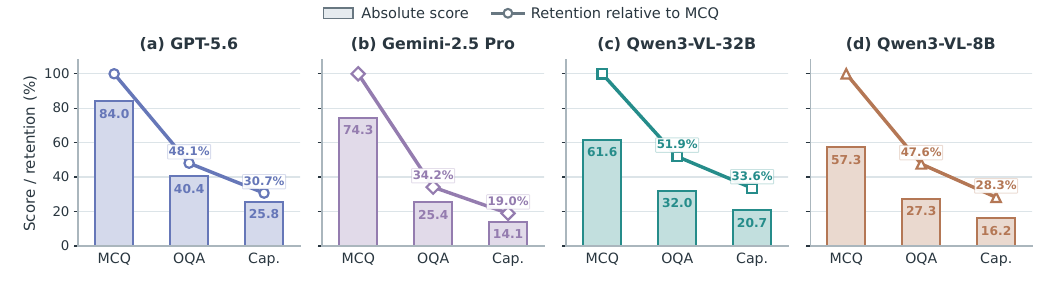}
    \caption{Anchor-fading profiles of representative Video-MLLMs on DualityVidQA-test (\textit{Both}) under our evaluation protocol. Each panel shows one model; bars report absolute scores and lines report retention relative to MCQ. The consistent decline from MCQ to OQA and captioning reveals strong reliance on instance-specific textual anchors.}
    \label{motivation}
\end{figure*}

Counterfactual video understanding investigates whether video-based multimodal large language models (Video-MLLMs) capture the physical and commonsense rules of dynamic environments \cite{huang2026taming,chen2025countervqa,motamed2025travl}. However, most existing benchmarks rely on multiple-choice questions (MCQ) \cite{huang2026taming,bai2025impossible}. This formulation provides strong textual priors. 
The question and candidate options consequently act as instance-specific textual anchors that specify the event to inspect, turning an inherently open-ended discovery problem into text-guided visual verification. High MCQ accuracy may therefore overestimate counterfactual understanding: a model can confirm a supplied hypothesis without autonomously determining what is anomalous. 
In this paper, we ask a simple question: \emph{Can a Video-MLLM independently discover, localize, and explain a counterfactual event without being told what to look for?}


To expose this overestimation, we progressively reduce instance-specific textual guidance from MCQ with a question and candidate options, to OQA with the question alone, and finally to captioning with only a generic instruction, as illustrated in Figure~\ref{fig:teaser}(a). Existing Video-MLLMs deteriorate sharply along this progression. On DualityVidQA-test \cite{huang2026taming}, GPT-5.6 \cite{singh2025openai} falls from 84.0 in MCQ to 40.4 in OQA and 25.8 in captioning, while Qwen3-VL-8B \cite{bai2025qwen3} drops from 57.3 to 27.3 and 16.2. Figure~\ref{motivation} reveals the same trend across representative open- and closed-source models, demonstrating that high MCQ scores can mask a shared dependence on text-specified hypotheses.

To bridge this gap, we introduce \textbf{FADE} (\textbf{F}ading Textual \textbf{A}nchors for Counterfactual \textbf{D}iscovery and \textbf{E}xplanation), an evidence-first, two-stage training framework that shifts counterfactual video understanding from text-guided verification toward video-driven discovery and explanation. In Stage I, evidence-internalized supervised fine-tuning (SFT) teaches the model to discover and temporally localize decisive counterfactual evidence. Building on this grounding, Stage II employs fading-anchor reinforcement learning (RL) to preserve evidence-based interpretation as questions and candidate options are progressively removed. As shown in Figure~\ref{fig:teaser}(b), training with FADE raises the OQA and captioning retention rates of Qwen3-VL-8B from 47.6\% and 28.3\% to 90.4\% and 67.4\%, respectively, with similarly substantial gains on IPV-Bench \cite{bai2025impossible}.


To evaluate this capability without constructing a new test set, we additionally develop an evaluation protocol that operates directly on existing public MCQ benchmarks. It preserves each video and its semantic target while reformulating the original query into three independently evaluated formats: MCQ with the question and candidate options, OQA with the question alone, and captioning with only a generic description instruction. Task-compatible verifiers then assess whether counterfactual judgments persist as instance-specific textual guidance is progressively removed.

Our contributions are threefold:
\begin{itemize}
    \item We identify textual anchoring as an under-explored problem in counterfactual video evaluation and distinguish passive verification from active discovery.
    \item We propose FADE, an evidence-first training framework that combines evidence-internalized SFT with fading-anchor RL to support autonomous counterfactual discovery and explanation.
    \item We develop a reusable evaluation protocol that reformulates MCQ instances from existing public benchmarks into progressively less guided formats without building a new benchmark from scratch. Extensive experiments expose the textual-anchor dependence of current Video-MLLMs and demonstrate the robustness of the FADE-trained model.
\end{itemize}

%% file: sections/2_relate.tex
\section{Related Works}
\subsection{Counterfactual Video Understanding}
Counterfactual video understanding probes whether models have genuinely acquired the real-world regularities underlying video events by exposing them to physical violations, causal interventions, or conflicts with commonsense knowledge. Early studies primarily followed the violation-of-expectation paradigm, assessing intuitive physical understanding by requiring models to distinguish physically plausible events from impossible ones \cite{riochet2018intphys,bordes2025intphys}.
Subsequent work, exemplified by CLEVRER and CoPhy, shifted the focus beyond event-plausibility judgment toward reasoning about temporal dynamics, causal dependencies, and counterfactual outcomes. This line of research was later extended to explaining violations of physical expectations and performing evidence-grounded and commonsense reasoning in complex real-world videos \cite{yi2019clevrer,baradel2019cophy,dai2023x,li2022representation}. 
With the emergence of multimodal large language models for video understanding, recent studies have introduced broader benchmarks covering diverse violations of real-world regularities and have improved anomalous-event reasoning through subproblem decomposition, paired real and counterfactual videos, and causal-graph-guided training \cite{zhou2025reasoning,bai2025impossible,huang2026taming,chen2025countervqa}. 
Nevertheless, most existing tasks provide hypotheses, targeted questions, or candidate answers that specify in advance which events the model should inspect and verify. Consequently, whether a model can move beyond explicit textual guidance to autonomously discover, localize, and explain anomalous content directly from video remains largely unexplored.

\subsection{Textual Bias in Multimodal Reasoning}
Multimodal models often exploit linguistic priors arising from question formulations, answer distributions, and textual co-occurrence patterns, rather than grounding their predictions in the visual input. Early VQA studies \cite{agrawal2018don,cadene2019rubi,niu2021counterfactual} exposed this vulnerability by modifying question-answer priors and subsequently sought to mitigate it through unimodal bias modeling and counterfactual causal inference, thereby reducing the direct influence of question semantics on answer prediction. This line of inquiry later expanded to video understanding. Through complementary diagnostic techniques, including visual evidence localization, modality ablation, and minimal video pairs, prior work \cite{li2022invariant, xiao2024can, krojer2025shortcut,lim2026video} has shown that high question-answering accuracy does not necessarily imply reliable video understanding: model predictions may still be driven by linguistic correlations, static visual cues, or temporally irrelevant segments rather than the evidence needed to support the answer. Moreover, TemporalBench \cite{cai2024temporalbench} and recent analyses of bias in multiple-choice video question answering \cite{loginova2025addressing} suggest that linguistic regularities and positional imbalances among answer options can introduce additional answer-selection shortcuts. Nevertheless, existing studies have largely examined whether models can ignore visual inputs or exploit dataset-level statistical biases. Much less attention has been paid to a subtler form of textual anchoring: even when a model processes the video, the question and candidate answers may predetermine what it searches for, turning open-ended anomaly discovery into the passive verification of an explicitly suggested phenomenon.

\subsection{Visual Grounding and Reinforcement Learning in Video MLLMs}
Visual grounding aligns model predictions with relevant video segments, thereby grounding video reasoning in verifiable spatiotemporal evidence. NExT-GQA \cite{xiao2024can} pioneered the joint evaluation of answer prediction and temporal evidence localization, while more recent approaches, such as Grounded-VideoLLM \cite{wang2024grounded} and ED-VTG \cite{pramanick2025enrich}, strengthen fine-grained temporal grounding in Video MLLMs through explicit temporal representations and query-semantic enrichment. Following the introduction of GRPO \cite{shao2024deepseekmath}, reinforcement learning has been increasingly adopted for video-model post-training. Video-R1~\cite{feng2026video} and VideoChat-R1~\cite{li2025videochat} improve video understanding by targeting temporal reasoning and multi-task spatiotemporal perception, respectively, whereas Time-R1~\cite{wang2026time} jointly optimizes reasoning and localization with verifiable temporal rewards. Related efforts further incorporate curriculum reinforcement learning to localize violation segments, as in RAVEN~\cite{ji2025raven}, or combine paired factual and counterfactual videos with a two-stage SFT--RL pipeline to mitigate language-prior dependence, as in DualityVidQA~\cite{huang2026taming}. Nevertheless, these methods typically optimize grounding or answer correctness under a fixed textual query, which specifies the event to be searched for in advance. In contrast, FADE progressively fades the guidance provided by questions and candidate answers, encouraging the model to move beyond query-conditioned evidence verification toward video-driven discovery and explanation of counterfactual events.

%% file: sections/3_method.tex
\section{FADE}
\label{sec:fade}
\begin{figure*}[htbp]
    \centering
    \includegraphics[width=0.9\linewidth]{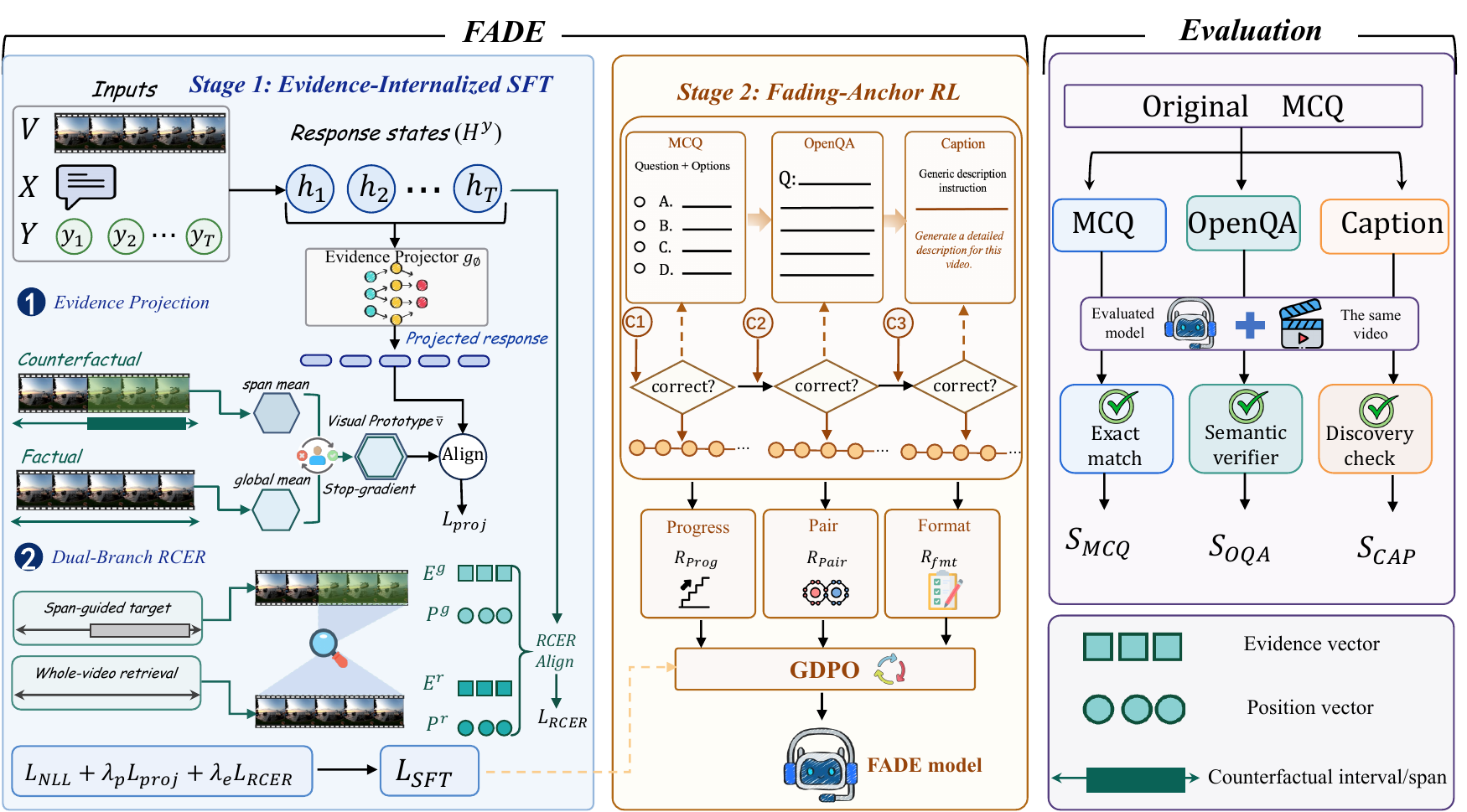}
\caption{Overview of FADE and the separate anchor-fading evaluation protocol. FADE comprises evidence-internalized SFT for discovering and grounding anomalous evidence (left) and fading-anchor RL for interpreting and explaining it under reduced textual guidance (center). Separately, the protocol reformulates each public MCQ benchmark instance into aligned MCQ, OQA, and captioning evaluations with task-compatible verifiers (right), requiring no new test set.}
\label{framework}
\end{figure*}


As illustrated in the left and center panels of Figure~\ref{framework}, FADE is an evidence-first, two-stage training framework. Stage I uses evidence-internalized supervised fine-tuning (SFT) to make decisive counterfactual evidence discoverable from the model's response representations. Building on this grounding, Stage II applies fading-anchor reinforcement learning (RL) to preserve evidence-based interpretation and explanation as instance-specific textual guidance is progressively reduced. The right panel summarizes the separate evaluation protocol described in Section \textbf{Experiments}; it is used only for evaluation and does not participate in FADE optimization.

\subsection{Stage I: Evidence-Internalized SFT}
\label{sec:sft}



Given a sample $(V_i,X_i,Y_i,z_i,\mathcal I_i)$, $V_i$ is the video, $X_i$ the
textual input, $Y_i=(y_{i,1},\ldots,y_{i,L_i})$ the target response, and
$z_i\in\{0,1\}$ the factual/counterfactual label, where $z_i=1$ denotes a
counterfactual video. For $z_i=1$, the normalized interval
$\mathcal I_i=[a_i,b_i]$ contains the decisive event; otherwise
$\mathcal I_i=\varnothing$. We retain the autoregressive negative
log-likelihood
\begin{equation}
    \mathcal L_{\mathrm{NLL}}
    =-\mathbb{E}_i\!\left[
      \frac{1}{L_i}\sum_{t=1}^{L_i}
      \log p_\theta(y_{i,t}\mid V_i,X_i,y_{i,<t})
    \right],
    \label{eq:fade_nll}
\end{equation}
where $p_\theta$ is the model distribution. This term preserves response correctness and fluency, but alone does not determine which visual evidence supports the response.



\paragraph{Localized evidence projection.}
As illustrated by the evidence-projection branch in the left panel of Figure~\ref{framework}, we use $\mathcal I_i$ only as privileged training supervision. Let $H_i^v=[\mathbf v_{i,1},\ldots,\mathbf v_{i,M_i}]$ be the temporally ordered visual tokens, $\tau_{i,m}$ the timestamp of $\mathbf v_{i,m}$, and $\mathcal S_i=\{m\mid a_i\leq\tau_{i,m}<b_i\}$. The target evidence prototype is
\begin{equation}
    \bar{\mathbf v}_i=
    \begin{cases}
      |\mathcal S_i|^{-1}\!\sum_{m\in\mathcal S_i}\mathbf v_{i,m}, & z_i=1,\\
      M_i^{-1}\!\sum_{m=1}^{M_i}\mathbf v_{i,m}, & z_i=0.
    \end{cases}
    \label{eq:evidence_prototype}
\end{equation}
Thus, counterfactual responses are tied to the annotated event, whereas a
global prototype avoids imposing an artificial anomaly location on factual
videos. For response states
$H_i^y=[\mathbf h_{i,1}^y,\ldots,\mathbf h_{i,L_i}^y]$, we define
\begin{equation}
    \mathcal L_{\mathrm{proj}}
    =\mathbb{E}_i\!\left[
      \frac{1}{L_i}\sum_{t=1}^{L_i}
      \left\|g_\phi(\mathbf h_{i,t}^y)
      -\operatorname{ng}(\bar{\mathbf v}_i)\right\|_2^2
    \right],
    \label{eq:evidence_projection}
\end{equation}
where $g_\phi$ is a lightweight projector and $\operatorname{ng}$ stops
gradients. This makes the decisive visual content recoverable from the
response representation rather than allowing a solution based only on textual
cues.

\paragraph{Response-conditioned evidence re-grounding.}
Evidence projection specifies what should support the response, but inference must recover that evidence without access to $\mathcal I_i$. The second blue branch in Figure~\ref{framework}, termed RCER, transfers interval supervision into whole-video retrieval. A span-guided branch first uses $K$ learnable queries $Q\in\mathbb{R}^{K\times d}$ to produce a privileged content-and-position target:
\begin{equation}
    (E_i^g,P_i^g)
    =\operatorname{Attn}
      (Q,H_{i,\mathcal S_i}^v,H_{i,\mathcal S_i}^v),
    \label{eq:span_guided_evidence}
\end{equation}
where $E_i^g\in\mathbb{R}^{K\times d}$ encodes evidence content and $P_i^g$
its temporal distribution. Factual videos instead use the global
representation and a uniform distribution. A response-conditioned branch,
which never observes $\mathcal I_i$, retrieves from the complete video:
\begin{equation}
    (E_i^r,P_i^r)
    =\operatorname{Attn}
      (W_q\widetilde H_i^y,H_i^v,H_i^v),
    \label{eq:response_conditioned_retrieval}
\end{equation}
where $\widetilde H_i^y$ contains $K$ sampled and projected response states,
and $W_q$ maps them into retrieval queries. We align the response states,
retrieved content, and temporal distributions through
\begin{equation}
\begin{aligned}
    \mathcal L_{\mathrm{RCER}}
    =\mathbb{E}_i\!\Bigl[
      &\alpha_1\|E_i^g-\widetilde H_i^y\|_1
      +\alpha_2\|E_i^r-\operatorname{sg}(E_i^g)\|_1 \\
      &+\alpha_3D_{\mathrm{KL}}
        \bigl(\operatorname{sg}(P_i^g)\|P_i^r\bigr)
    \Bigr].
\end{aligned}
    \label{eq:evidence_regrounding}
\end{equation}
The three terms respectively internalize the privileged evidence, recover its
content from the full video, and distill its temporal location. Stage I
optimizes
\begin{equation}
    \mathcal L_{\mathrm{SFT}}
    =\mathcal L_{\mathrm{NLL}}
    +\lambda_p\mathcal L_{\mathrm{proj}}
    +\lambda_e\mathcal L_{\mathrm{RCER}},
    \label{eq:fade_sft}
\end{equation}
where $\lambda_p$ and $\lambda_e$ balance the auxiliary objectives. Temporal
annotations and auxiliary branches are used only during training; the
evidence-internalized checkpoint initializes Stage II.

\subsection{Stage II: Fading-Anchor RL}
Although Stage I grounds the response in visual evidence, the textual input
may still reveal which event should be examined. The center panel of Figure~\ref{framework} therefore progressively broadens the
evidence-search space through three views of the same video:
\begin{equation}
    \mathcal A_i^{(1)}=(q_i,\mathcal O_i),\qquad
    \mathcal A_i^{(2)}=q_i,\qquad
    \mathcal A_i^{(3)}=\varnothing,
    \label{eq:anchor_hierarchy}
\end{equation}
corresponding to MCQ, OQA, and captioning. Here,
$\mathcal A_i^{(3)}=\varnothing$ removes all instance-specific anchors while
retaining a generic video-description instruction. We organize these views
into a scaffold-and-consolidate curriculum
\begin{equation}
    \mathcal C_1:1\Rightarrow2\Rightarrow3,\qquad
    \mathcal C_2:2\Rightarrow3,\qquad
    \mathcal C_3:3,
    \label{eq:fading_curriculum}
\end{equation}
where $\Rightarrow$ is a correctness-gated transition. $\mathcal C_1$ uses
stronger anchors to scaffold weaker-anchor behaviors; its checkpoint
initializes $\mathcal C_2$, and subsequently $\mathcal C_3$, each beginning
from a clean context. The model therefore first learns each harder behavior
and then consolidates it after the preceding textual cues are removed.

\paragraph{Trajectory rewards and policy optimization.}
Let $c_{i,k}^{(\ell)}\in\{0,1\}$ indicate whether trajectory $k$ for sample
$i$ passes the verifier at level $\ell$. MCQ uses exact matching, whereas OQA
and captioning use task-adapted semantic verification. For curriculum
$\mathcal C_r$, the prefix-gated progress reward is
\begin{equation}
    R_{i,k}^{\mathrm{prog}}
    =\sum_{\ell=r}^{3}\prod_{j=r}^{\ell}c_{i,k}^{(j)},
    \label{eq:progress_reward}
\end{equation}
so a weaker-anchor success is rewarded only if all preceding levels are also
correct.

To discourage an ``always counterfactual'' shortcut, each counterfactual video
$V_i^+$ is paired with its factual counterpart $V_i^-$. Let
$d_r(\widehat Y)\in\{0,1\}$ be the predicted direction under the starting
format of $\mathcal C_r$, where $1$ denotes counterfactual. We define
\begin{equation}
    R_{i,k}^{\mathrm{pair}}
    =\mathbb{I}\!\left[
      d_r(\widehat Y_{i,k}^{+})=1
      \ \land\
      d_r(\widehat Y_{i,k}^{-})=0
    \right],
    \label{eq:pair_reward}
\end{equation}
which is positive only when both directions are correct. A lightweight
$R_{i,k}^{\mathrm{fmt}}$ additionally ensures parseable responses.

Because the progress, pair, and format signals differ in scale and sparsity,
we adopt GDPO~\cite{liu2026gdpo} to normalize them independently. For $G$
trajectories sampled from prompt $i$, the advantage is
\begin{equation}
    A_{i,k}
    =\operatorname{Norm}_{\mathcal B}\!\left[
      \sum_{c\in\{\mathrm{prog,pair,fmt}\}}w_c
      \frac{R_{i,k}^{c}-\mu_i^{c}}
           {\sigma_i^{c}+\epsilon}
    \right],
    \label{eq:gdpo_advantage}
\end{equation}
where $\mu_i^c$ and $\sigma_i^c$ are the group statistics of component $c$,
$w_c$ is its weight, and $\operatorname{Norm}_{\mathcal B}$ denotes batch
whitening. We then apply the standard KL-regularized clipped objective with
the Stage-I model as the reference policy. Thus, SFT establishes discoverable
visual evidence, while RL preserves its use as textual anchors fade; neither
temporal annotations nor paired videos are required at inference.

%% file: sections/4_experiement.tex
\section{Experiments}
\begin{table*}[h]
    \centering
    \small

  \begin{adjustbox}{max width=\linewidth} 

  \begin{tabular}{lcccccccccccc}
\hline
                        & \multicolumn{9}{c}{DualityVidQA}                                                                                                              & \multicolumn{3}{c}{IPV-Bench}                 \\ \cline{2-13} 
                        & \multicolumn{3}{c}{MCQ}                       & \multicolumn{3}{c}{OQA}                   & \multicolumn{3}{c}{CAP.}                   & MCQ           & OQA       & CAP.       \\ \cline{2-13} 
\multirow{-3}{*}{Model} & Real          & CF            & Both          & Real          & CF            & Both          & Real          & CF            & Both          & /             & /             & /             \\ \hline
\rowcolor[HTML]{FFFFFF} 
Random                  & 24.2          & 23.9          & 4.5           & /             & /             & /             & /             & /             & /             & 20            & /             & /             \\ \hline
\rowcolor[HTML]{C0C0C0} 
\multicolumn{13}{c}{\cellcolor[HTML]{C0C0C0}\textit{\textbf{Close-source VLLMs}}}                                                                                                                                       \\ \hline
GPT-4o-mini \cite{hurst2024gpt}             & 91.8          & 57.4          & 51.9          & 91.7          & 26.7          & 24.7          & 93.2          & 17.7          & 11.3          & 68.4          & 25.6          & 14.1          \\
GPT-4o \cite{hurst2024gpt}                 & 92.7          & 72.1          & 65.8          & 93.0          & 27.4          & 21.8          & 94.5          & 21.1          & 14.6          & 79.7          & 35.3          & 21.8          \\
GPT-4.1 \cite{achiam2023gpt}                 & 87.3          & 75.5          & 65.6          & 90.5          & 25.1          & 23.4          & 91.2          & 16.3          & 12.2          & 82.6          & 35.6          & 21.3          \\
GPT-5.5 \cite{openai2026gpt55}                 & 95.0          & 83.4          & 82.4         & 96.5          & 38.6          & 35.5          & 97.0          & 24.8          & 22.1          & 89.5          & 40.2          & 24.9          \\
GPT-5.6 \cite{openai2026gpt56} & \textbf{96.8} & 85.5 & 84.0 & 98.2 & 40.7 & 40.4 & 99.0 & 25.8 & 25.8 & 90.6 & 43.1 & 25.5 \\
Gemini-2.5 Pro \cite{comanici2025gemini}          & 92.5          & 80.3          & 74.3          & 94.3          & 31.8          & 25.4          & 94.5          & 16.1          & 14.1          & 87.4          & 39.1          & 24.2          \\ \hline
\rowcolor[HTML]{C0C0C0} 
\multicolumn{13}{c}{\cellcolor[HTML]{C0C0C0}\textit{\textbf{Open-source VLLMs}}}                                                                                                                                        \\ \hline
Qwen2.5-vl-7B$^*$ \cite{bai2025qwen2}          & 91.7          & 59.9          & 52.8          & 94.2          & 21.5          & 14.9          & 92.8          & 12.1          & 9.8           & 73.2          & 26.3          & 14.4          \\
Qwen2.5-vl-32B$^*$ \cite{bai2025qwen2}        & 95.2          & 55.4          & 51.3          & 96.8          & 32.3          & 29.3          & 97.8          & 12.1          & 11.4          & 77.6          & 29.9          & 17.1          \\
Qwen2.5-vl-72B$^*$ \cite{bai2025qwen2}          & 95.8          & 62.8          & 58.9          & 97.8          & 35.7          & 35.2          & 96.8          & 15.6          & 13.7          & 81.3          & 34.1          & 20.3          \\
Qwen3-vl-8B \cite{bai2025qwen3}             & 91.7          & 64.4          & 57.3          & 91.1          & 28.7          & 27.3          & 94.3          & 18.0          & 16.2          & 76.8          & 29.1          & 16.1          \\
Qwen3-vl-32B \cite{bai2025qwen3}            & 93.3          & 70.5          & 61.6          & 93.8          & 33.9          & 32.0          & 95.7          & 20.9          & 20.7          & 83.5          & 36.0          & 22.0          \\
InternVL3.5-8B \cite{wang2025internvl3}         & 39.1          & 77.3          & 18.9          & 40.4          & 21.5          & 7.8           & 77.3          & 9.4           & 9.4           & 74.6          & 26.7          & 14.7          \\
InternVL3.5-14B \cite{wang2025internvl3}         & 36.4          & 81.1          & 19.7          & 35.7          & 21.7          & 8.4           & 81.3          & 5.3           & 5.3           & 78.2          & 31.1          & 17.7          \\
MiMo-VL-7B \cite{team2025kimi}              & 74.6          & 76.1          & 52.6          & 81.4          & 14.0          & 11.2          & 91.3          & 11.7          & 8.0           & 75.1          & 26.9          & 14.4          \\
DNA-Training-7B$^*$ \cite{huang2026taming} & 95.8 & 80.1 & 76.8 & 96.8 & 48.7 & 46.2 & 97.5 & 28.0 & 25.8 & 85.7 & 43.8 & 25.2 \\\hline
FADE(ours)                    & 96.4 & \textbf{87.7} & \textbf{84.6} & \textbf{99.2} & \textbf{77.8} & \textbf{76.5} & \textbf{99.8} & \textbf{57.2} & \textbf{57.0} & \textbf{91.2} & \textbf{78.6} & \textbf{60.2} \\ \hline
\end{tabular}

\end{adjustbox}
\caption{Results (\%) under our anchor-fading evaluation protocol on
DualityVidQA-test and IPV-Bench. For DualityVidQA-test, Real, CF, and Both denote factual, counterfactual, and strict paired accuracy, respectively; IPV-Bench contains only counterfactual videos. For models marked with $^{*}$, MCQ scores are taken from the original DualityVidQA study \cite{huang2026taming}; all other scores are obtained using our protocol.}
\label{main_result}

\end{table*}

\begin{figure*}[htbp]
    \centering
    \includegraphics[width=\linewidth]{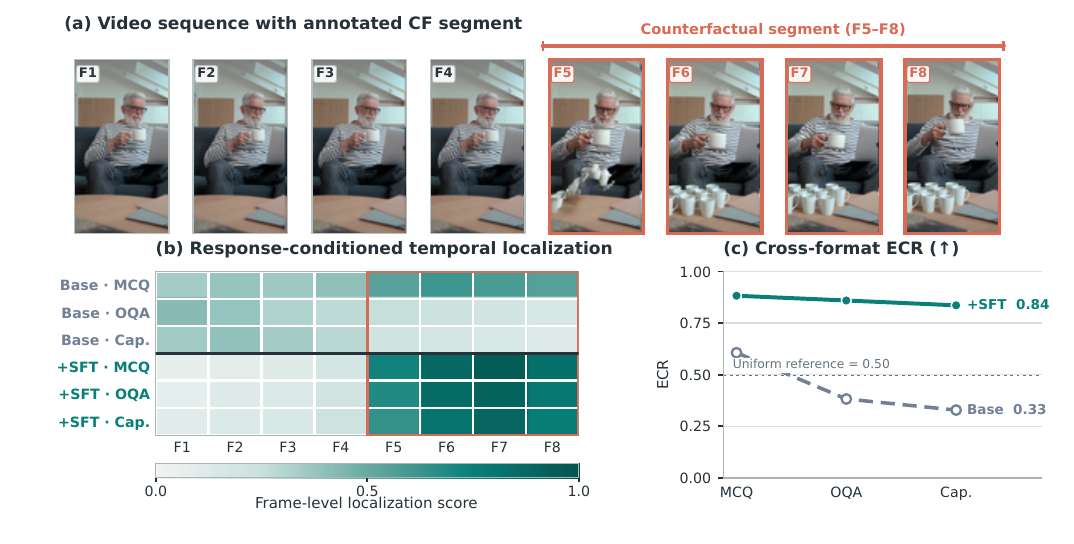}
\caption{Visualization of counterfactual evidence discovery before and after
SFT. (a) Video sequence with the annotated counterfactual interval (F5--F8).
(b) Frame-level localization responses of the base and SFT-tuned models across
MCQ, OQA, and captioning. (c) Evidence concentration rate (ECR), showing that
SFT consistently focuses on decisive evidence as textual anchors fade.}
\label{fig:sft_localization}
\end{figure*}

\subsection{Evaluation Protocol}
\label{sec:anchor_fading_eval}

Existing counterfactual-video benchmarks are predominantly formulated as MCQs, whose questions and candidate options specify both what to inspect and which hypotheses to verify. To diagnose this textual anchoring without collecting a new test set, we reformulate each MCQ instance from an existing public benchmark into three semantically aligned evaluations with progressively less instance-specific guidance, as shown in the right panel of Figure~\ref{framework}.

Given a source instance
$\mathcal{B}_i=(V_i,q_i,\mathcal{O}_i,a_i^\star)$, let
$e_i^\star$ denote the target-event semantics jointly specified
by $(q_i,o_{i,a_i^\star})$. We construct
\begin{equation}
\begin{aligned}
\mathcal{U}_i^{\mathrm{MCQ}} &= (V_i,q_i,\mathcal{O}_i),\\
\mathcal{U}_i^{\mathrm{OQA}} &= (V_i,q_i),\\
\mathcal{U}_i^{\mathrm{CAP}} &= (V_i,p_{\mathrm{cap}}).
\end{aligned}
\end{equation}

where $p_{\mathrm{cap}}$ is a generic description instruction shared across all samples. Thus, instance-specific anchors fade as
$(q_i,\mathcal{O}_i)\rightarrow q_i\rightarrow\emptyset$. The three formats preserve the original video, target semantics, public split, and visual sampling budget, and are evaluated independently from clean contexts.

For an arbitrary evaluated model $f_{\psi}$, correctness and the reported score at level $\ell\in\{\mathrm{MCQ},\mathrm{OQA},\mathrm{CAP}\}$ are
\begin{equation}
\begin{aligned}
    c_i^{(\ell)}
    &=\mathcal{J}_{\ell}\!\left(
f_{\psi}(\mathcal{U}_i^{(\ell)});
a_i^\star,e_i^\star,q_i
\right)\in\{0,1\},\\
    S^{(\ell)}
    &=\frac{1}{N}\sum_{i=1}^{N}c_i^{(\ell)}.
\end{aligned}
\label{eq:eval_scores}
\end{equation}
Here, $\mathcal{J}_{\mathrm{MCQ}}$ uses exact option matching, $\mathcal{J}_{\mathrm{OQA}}$ uses a frozen semantic verifier, and $\mathcal{J}_{\mathrm{CAP}}$ checks whether the target event is independently discovered and correctly described. The resulting profile $\mathbf{F}_{\psi}=(S^{(\mathrm{MCQ})},S^{(\mathrm{OQA})}, S^{(\mathrm{CAP})})$ diagnoses whether video-grounded judgments persist as textual anchors fade. Because the output space also becomes increasingly open-ended, cross-format gaps are interpreted diagnostically rather than as strict causal effects.

\subsection{Experimental Setup}


\paragraph{Training Data.}
We directly use the DualityVidQA \cite{huang2026taming} training split. The SFT data contain 104,879 samples, comprising 54,879 factual and 50,000 counterfactual videos, while the RL data contain 20,000 factual--counterfactual video pairs. Because the original training annotations are MCQ-only, we supplement them with OQA answers and captions generated by Qwen3.6-Plus \cite{qwen36plus} using each question and correct option as semantic context. These references support FADE's fading-anchor curriculum while preserving the original MCQ supervision.

\paragraph{Implementation Details.} We adopt Qwen3-VL-8B \cite{bai2025qwen3} as the base model and implement the SFT and RL phases using the \texttt{LLaMA-Factory} \cite{zheng2024llamafactory} and \texttt{ms-swift} \cite{zhao2024swiftascalablelightweightinfrastructure} frameworks, respectively. The SFT phase is fine-tuned for two epochs using LoRA with a rank of 8, with a learning rate of $5 \times 10^{-5}$. The RL phase is initialized from the SFT checkpoint, trained for one epoch with a learning rate of $1 \times 10^{-6}$, and samples 8 candidate responses per input. For both stages, we set the per-GPU batch size to 1, apply gradient accumulation over 4 steps, and utilize BF16 precision. All experiments were conducted on a cluster equipped with 8 NVIDIA H20 GPUs.

\paragraph{Benchmarks and Metrics.}
We apply the evaluation protocol to DualityVidQA-test \cite{huang2026taming} and IPV-Bench \cite{bai2025impossible}. DualityVidQA-test contains factual--counterfactual video pairs and reports factual accuracy (Real), counterfactual accuracy (CF), and strict paired accuracy (Both), where a pair is correct only when both instances are correctly interpreted. IPV-Bench contains only counterfactual videos and reports accuracy under each reconstructed format. Thus, the protocol reuses the original public test instances while diagnosing whether model judgments survive the removal of textual guidance.

\subsection{Main Results}

Table~\ref{main_result} reports results with our evaluation protocol. Existing Video-MLLMs deteriorate sharply as textual guidance is removed. On DualityVidQA-test (\textit{Both}), Qwen3-VL-8B drops from 57.3 in MCQ to 27.3 in OQA and 16.2 in captioning, retaining only 47.6\% and 28.3\% of its MCQ score. Even GPT-5.6 falls from 84.0 to 40.4 and 25.8, with retention rates of 48.1\% and 30.7\%. These consistent declines indicate that strong MCQ performance can reflect verification of text-specified hypotheses rather than autonomous counterfactual discovery.

The FADE-trained model achieves the best strict paired scores on DualityVidQA-test and the best scores on IPV-Bench across all three formats. On DualityVidQA-test (\textit{Both}), it scores 84.6, 76.5, and 57.0 under MCQ, OQA, and captioning, retaining 90.4\% and 67.4\% of its MCQ performance. On IPV-Bench, it obtains 91.2, 78.6, and 60.2, compared with 90.6, 43.1, and 25.5 for GPT-5.6. The substantially flatter degradation demonstrates that FADE-trained model preserves evidence-grounded counterfactual judgments as instance-specific textual anchors fade.

\subsection{Ablation Study}

\begin{table}[t]
    \centering
    \small
    \scriptsize
    \setlength{\tabcolsep}{2.4pt}
    \resizebox{\columnwidth}{!}{
    \begin{tabular}{lcccccc}
        \toprule
        & \multicolumn{3}{c}{DualityVidQA-test} 
        & \multicolumn{3}{c}{IPV-Bench} \\
        \cmidrule(lr){2-4}\cmidrule(lr){5-7}
        Configuration & MCQ & OQA & Cap. & MCQ & OQA & Cap. \\
        \midrule
        Qwen3-VL-8B   & 57.3 & 27.3 & 16.2 & 76.8 & 29.1 & 16.1 \\
        w/o SFT       & 63.8 & 37.9 & 24.6 & 80.2 & 40.5 & 26.7 \\
        w/o RL        & 77.9 & 55.6 & 37.4 & 85.9 & 57.2 & 38.7 \\
        w/o Prog. RL  & 82.1 & 67.8 & 48.8 & 89.1 & 69.5 & 50.8 \\
        \midrule
        FADE          & \textbf{84.6} & \textbf{76.5} & \textbf{57.0}
                      & \textbf{91.2} & \textbf{78.6} & \textbf{60.2} \\
        \bottomrule
    \end{tabular}
    }
    \caption{Ablation of FADE on DualityVidQA-test (\textit{Both}) and IPV-Bench under the same evaluation protocol, showing the complementary contributions of SFT, RL, and progressive anchor fading.}
\label{ablation}
\end{table}

\paragraph{From Evidence Discovery to Understanding.}
To investigate the individual contributions of SFT, RL, and the progressive reinforcement strategy, we construct three ablation variants: \textit{w/o SFT}, which applies progressive RL directly to the base model; \textit{w/o RL}, which retains only the SFT phase; and \textit{w/o Prog. RL}, which jointly trains on MCQ, OpenQA, and Captioning with identical data and update steps to ablate the progressive mechanism. All variants share the same base model and evaluation protocol. For DualityVidQA-test, we exclusively report the strict \texttt{Both} metric.

As shown in Table~\ref{ablation}, directly applying RL yields only marginal gains, indicating that in the absence of evidence-aware initialization, the model struggles to consistently localize critical counterfactual segments. While the SFT-only variant significantly improves MCQ performance, it exhibits a substantial deficit on openQA and captioning, suggesting that evidence retrieval alone is insufficient to guarantee accurate comprehension. The non-progressive RL variant further enhances overall performance; however, its advantages remain largely contingent on scenarios with strong textual anchors. As questions and options are progressively stripped of textual cues, its performance gap relative to the full model consistently widens. The full FADE-trained model achieves the best performance across all configurations on both benchmarks, demonstrating that evidence discovery (via SFT), evidence comprehension (via RL), and robustness to weak textual cues (via progressive optimization) are mutually indispensable.

\paragraph{Counterfactual Evidence Discovery Capability of SFT.} To investigate whether SFT effectively equips the model with the ability to identify definitive counterfactual evidence, Figure~\ref{fig:sft_localization} presents a comparison of per-frame evidence responses between the base model and an SFT-only variant across three conditions: MCQ, OpenQA, and Caption. We quantify the concentration of model responses within the ground-truth counterfactual interval $\mathcal{T}_{\mathrm{CF}}$ using the Evidence Concentration Rate (ECR), defined as $\mathrm{ECR}=\sum_{t\in\mathcal{T}_{\mathrm{CF}}}s_t/\sum_t s_t$, where $s_t$ denotes the localization score at frame $t$. The base model accurately localizes counterfactual segments only under the MCQ condition, where explicit answer options are provided; the progressive removal of textual cues causes its attention distribution to become increasingly diffuse. In contrast, the SFT-tuned model consistently localizes the counterfactual intervals across all three conditions. This demonstrates that SFT shifts the model from prompt-dependent passive verification to autonomous, task-agnostic evidence discovery, thereby validating the primary design objective of Phase~I: enabling the model to accurately attend to pivotal evidence.

%% file: sections/5_conclusion.tex
\section{Conclusion}

We present FADE, a two-stage training framework that shifts counterfactual video understanding from text-guided verification toward video-driven discovery. Evidence-internalized SFT first grounds model responses in decisive anomalous evidence, and fading-anchor RL then preserves its interpretation and explanation as textual guidance decreases. In addition, we introduce an anchor-fading evaluation protocol that reformulates existing public MCQ benchmarks into aligned MCQ, OQA, and captioning settings without collecting a new test set. Experiments on DualityVidQA and IPV-Bench reveal that existing Video-MLLMs degrade sharply as textual anchors fade, whereas the FADE-trained model remains substantially more robust, underscoring the importance of evaluating autonomous evidence discovery beyond constrained MCQ accuracy.